# Hybrid Technique for effective Knowledge representation & a Comparative Study

**Poonam Tanwar [1], Dr. T. V. Prasad[2], Dr. Kamlesh Datta[3]**

[1] Asst. Professor, Dept. of CSE, Lingaya's University, Faridabad, Hryana, India & PhD Scholar, Uttarakhand Technical University, Dehradun, Uttarakhand, India
`poonam.tanwar@rediffmail.com`

[2] Dean (R&D), Lingaya's University, Faridabad, Haryana, India
`tvprasad2002@yahoo.com`

[3] Associate Prof & HOD (CSE), National Institute of Technology, Hamirpur, Himachal Pradesh, India
`kdnith@gmail.co m`

## ABSTRACT

*Knowledge representation(KR) and inference mechanism are most desirable thing to make the system intelligent. System is known to an intelligent if its intelligence is equivalent to the intelligence of human being for a particular domain or general. Because of incomplete ambiguous and uncertain information the task of making intelligent system is very difficult. The objective of this paper is to present the hybrid KR technique for making the system effective & Optimistic. The requirement for (effective & optimistic) is because the system must be able to reply the answer with a confidence of some factor. This paper also presents the comparison between various hybrid KR techniques with the proposed one.*

.

## 1. INTRODUCTION

1 Knowledge representation
Knowledge representation: - In AI system implementation, efficiency, speed and maintenance are the major things affected by the knowledge representation. A KB structure must be capable of representing the broad spectrum of knowledge types categorized by Feigenbaum include [5].

- Objects - information on physical objects and concepts
- Events - time-dependent actions and events that may indicate cause and effect relationships.
- Performance – procedure or process of performing tasks
- Meta-knowledge – knowledge about knowledge including its reliability, importance, performance evaluation of cognitive processors.

Many of the problems in AI require extensive knowledge about the world. Objects, properties, categories and relations between objects, situations, events, states and time, causes and effects are the things that AI needs to represents. Knowledge representation provides the way to represent all the above defined things [38]. KR techniques are divided in to two major categories that are Declarative representation & Procedural representation. The Declarative representation techniques are used to represents objects, facts, relations. Whereas the Procedural representation are used to represent the action performed by the objects. The propositional logic, predicate logic, semantic net are the declarative knowledge representation techniques and Script, Conceptual dependency are procedural knowledge representation techniques. There is one more technique named frame that can be used as both. Each one has their own prone and cons. Now days because of market demands there are number of hybrid techniques are available. In next section we cover the few hybrid KR techniques.

## 1.1 KRYPTON

In 1983 Ronald J. Brachman, Richard E. Fikes, Hector J. Levesque has developed krypton a hybrid knowledge representation technique.

**Technical description:** Two boxes are used terminological box(T box) and assertion box (A box). TBox used the structure of KL-ONE in which terms are organized taxonomically, using frames an ABox used the first-order logic sentences for those predicates come from the TBox, and a symbol table maintaining the names of the TBox terms so that a user can refer to them. It is basically a tell ask module. All interactions between a user and a KRYPTON knowledge base are mediated by TELL and ASK operations shown in fig 3.1[14][16].

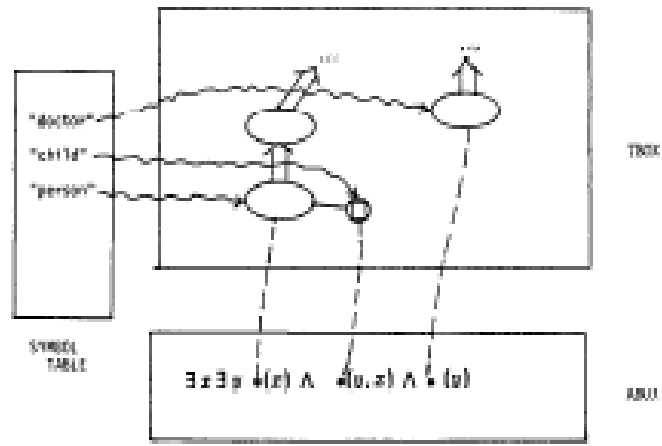

Fig.1 Overview of KRYPTON [14]

## 1.2 OBLOG 2:-

In 1987, Thhomas F. Gordon Presented the Oblog 2. Oblog stand for Object-oriented LOGic, is an experimental hybrid knowledge representation and reasoning system. It is a hybrid of a terminologicalreasoner with a Prolog inference mechanism. The description of type and attribute taxonomies are supported by terminological component whereas Entities are instances of a set of types. Horn clause rules are used as Procedures, & for determining the values of attributes are indexed by type.

## 1.3 MANTRA

In 1991 MANTRA was developed by J. Calmet, I.A. Tjandra, G. Bittencourt.

**Technical description:** It is the combination of four different knowledge representation techniques. First order logic, terminological language, semantic networks and Production systems. all algorithm used for inference are decidable because this representation used the four value logic. Mantra is a three layers architecture model. It consist the epistemological level, the logical level, Heuristic level.

## 1.4 FRORL:

(frame-and-rule oriented requirement specification Language) was developed by Jeffrey J. P. Tsai, Thomas Weigert, Hung-Chin Jang in 1992.

**Technical Description:** FRORL is based on the concepts of frames and production rules Which is designed for software requirement and specification analysis. There are two types of frames Object frame and activity frames. Object Frames are used to represent the real world entity not limited to physical entity. These are act as a data structure. Each activity in FRORL is represented by activity frame to represent the changes in the world. Activity, Precondition and action are reserved word not to be used in specification. FRORL consist of Horn clause of predicate logic. The comparisons between various hybrid KR techniques with the proposed technique are shown in table 1.

## 2. KNOWLEDGE BASE SYSTEM

The KR system must be able to represent any type of knowledge, "Syntactic, Semantic, logical, Presupposition, Understanding ill formed input, Ellipsis, Case Constraints, Vagueness". In our previous paper we have proposed the model for effective knowledge representation technique that consist five different parts the K Box, Knowledge Base, Query applier, reasoning and user interface as shown in fig 8. This time the total emphasis is on knowledge representation. This section used to describe the new hybrid knowledge representation technique which is the integration of script and semantic net KR technique. The semantic net & script KR technique are explained in next subsection.

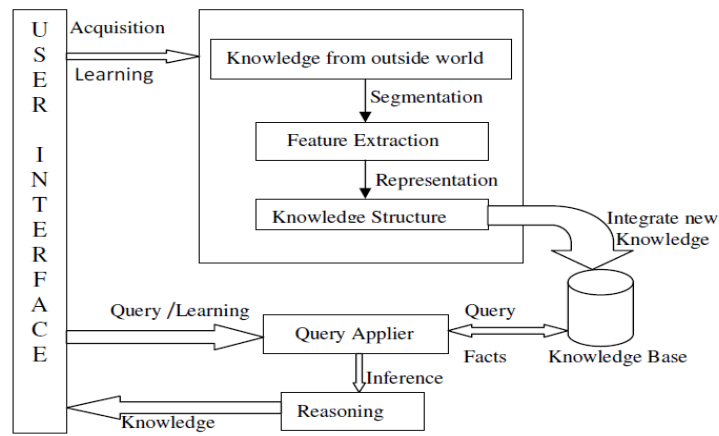

Fig 2. Knowledge Base System Model /Architecture [39].

## 2.1 SEMANTIC NET

A semantic network is widely used knowledge representation technique. Semantic network is a KR technique in which the relationship between class and objects are represented by the connection/link between objects or class of objects.

The nodes / vertices in semantic net are used to represent the Generic class or a particular class or an instance of a class (object).Relation between them is represented by the link, which shows the activation comes from where .The links are unidirectional .these links represents the semantic relationship between the objects. Semantic network are generally used to represent the inheritable knowledge. Inheritance is most useful form of inference. Inheritance is the belongings in which element of some class inherit the attribute and values from some other class shown in Fig.1 [38].

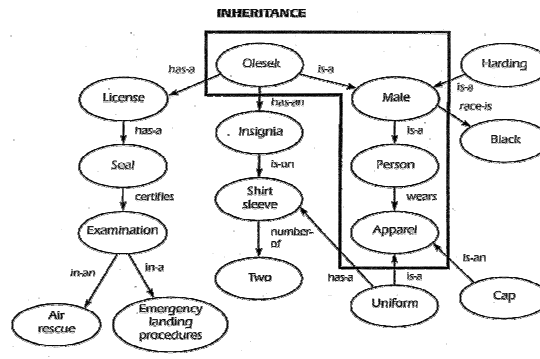

Fig.3 represents the inheritance relation [35][38].

Because there is an association between two or more nodes the Semantic nets are also known as associative nets. These associations are proved to be useful for inferring some knowledge from the existing one. If user wants to get any knowledge from the knowledge base they need not to put any query. The activated association or relation provides the result directly or indirectly only need to follow the links in the semantic net. IS-A, and A-KIND-OF are generally used to represent the value of a link in semantic net shown in fig 2.

KR techniques are divided in to two main categories one is declarative and other is procedural. Semantic net is a declarative KR technique that can be used either to represent knowledge or to support automated systems for reasoning about knowledge. Semantic net can be used in variety of ways, as per the requirement following are six of the most common kinds of semantic networks.

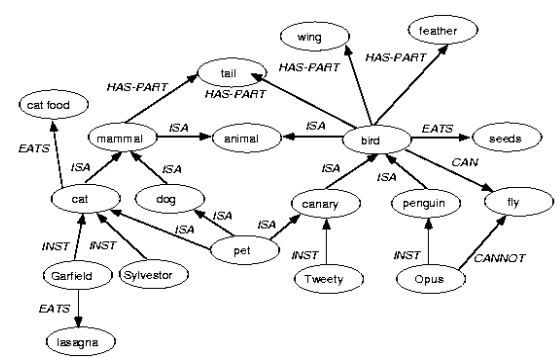

**Fig 4 Represents of IS-A, HAS, INSTANCE [17], [38].**

1. Definitional networks
2. Assertional networks
3. Implicational networks
4. Executable networks
5. Learning networks
6. Hybrid networks

During 1975 (See Walker ) Partitioned semantic net came in picture for speech understanding system. Then after that in 1977 Hendrix explained how we can expend the utility of semantic net using partitioned semantic net [8].In case of a huge network  semantic net can be divided in to two  more net. The semantic net is to be partitioned to separate the various nodes and arcs in to units and each unit is known as spaces. Using partitioned semantic net user can define the existence of the entity. One space is assigned to every node and arc and all nodes and arcs lying in the same space are distinguishable from those of other spaces. Nodes and arcs of different spaces may be linked, but the linkage must pass through the boundaries which separate one space from another [38].

Partitioning semantic nets can be used to delimit the scopes of quantified variables. While working with quantified statements, it will be help full to represent the   pieces of information consist some event .For ex "Poonam believes that earth is round " is represented by the fig 3. Nodes<POONAM>' is an agent of Event node.<EARTH>' and <ROUND> represent the objects of space1.

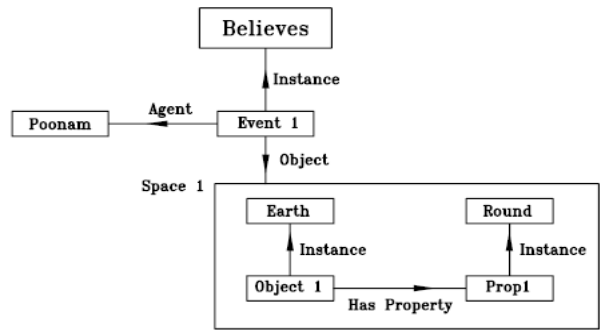

Fig.5 Partitioned Semantic Net [38]

Universal and existential quantifier can also be represent by the Partitioning semantic net. For ex, "Every sister knots the rakhee to her brother" in predicate logic. In predicate logic the sister S and rakhee R are represented as objects while the knot event is expressed by a predicate where as in case of semantic net the event is represented as an object of some complex object, i.e., the bite event is a situation which could be the object of some more complex event. Partitioning semantic net can also be used to represent universal quantifier. For ex "Every sister knots the rakhee to her brother" is represented in fig 4 [38]. Partitioning semantic net can also be used for complex quantifleations which involve nested scopes by using nesting space.

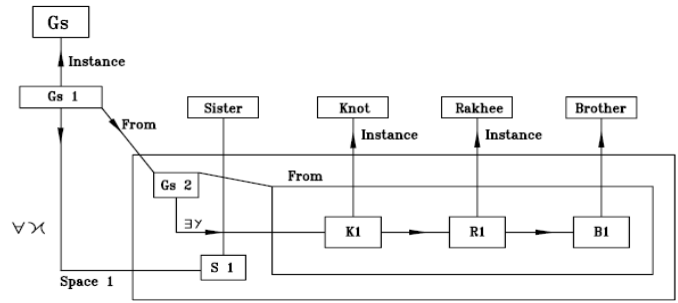

Fig.6 Represents Partitioned Semantic Net for Quantifiers [38].

## 2.2 Scripts

A variation in the theme of structured objects called scripts was devised by Roger Schank and his associates in 1973[3].It is an active type information which contain class of events in terms of contexts, participants and sub-events represented in the form of collection of slots or series of frames which uses inheritance and slots . Scripts predict unobserved events and can build coherent account from disjointed observations. Scripts basically describes the stereotypical knowledge i.e if the system in not given the information dynamically then it assumes the default information to be true Scripts are beneficial because real world events do follow stereotyped patterns as human beings use previous experiences to understand verbal accounts. A script is used for organizing the knowledge as it directs the attention and recalls the inference. They provide knowledge and expectations about specific events or experiences and can be applied to new situations. For example: "Rohan went to the restaurant and had some pastries". it was good now meaning derived from the above text one gets to know he got the pastries from the restaurant and that for eating and that was good. Script defines an episode with the known behavior and describes the sequence of events. The script consist the following.
- Current plans (Entry condition, Result)
- Social link(Track)
- Played roles,
- Scene.
- Probs.
- Anything indicating the behavior of the script in a given situation.

An example of script for class room is shown in fig.7.

| Script Lecture Room | |
|---|---|
| Track: Class Room | Entry Cond: T has prepared lecture. |
| Props: Table, Chair, Chock Board, Chock Box, Duster, Lecture Stand, Projector. | T has Lecture Notes. The class is open. T has attendance register. |
| Roles:   T = Teacher          S= Student | Result :   T has imparted knowledge.          S : Acquired Knowledge. |

| Script Lecturer Room Contd. | |
|---|---|
| **Scene 1 ENTERING**<br><br>T : enter the classroom.<br>T : moves to lecture stand.<br>T: switch on the projector.<br>T: Look the student. | **Scene 2 LECTURE**<br><br>T : Lecture notes on lecture stand<br>T : Select the lecture no.<br>T : Explain the lecture.<br><br>S: Listen the lecture.<br>S: ask the question.<br>T : use the board.<br>T : go to the scene 4 at the "No Student in class"<br>T : Explain.<br><br>T: Ask the question. |

| Script Lecturer Room Contd. | |
|---|---|
| **Scene 3 Question Solving**<br><br>T: gave question.<br>S : discussion.<br>S: Solve the question.<br>T: Solve the question. | **Scene 4 Exiting**<br><br>T : Took the attendance.<br>T : Collect the sheet.<br>T : Leave the class room. |

Fig.7 Script structure for class room.

Advantages of using scripts:
- Details for a particular object remain open and
- Reduces the search space.

Disadvantages
- Less general than Frames
- It may not be suitable for all kind of Knowledge

## 3 HYBRID KNOWLEDGE REPRESENTATION TECHNIQUE

Every knowledge representation technique has their own merits and demerits that depend on which type of knowledge we want to represent. To navigate the problem associated with single knowledge representation technique the hybrid knowledge representation came in picture
The script and semantic net alone is a strong representation technique but still they have some disadvantages. The previous section consist the example of script for lecture room using that we are unable to get the detail like the teacher can teach one or more subject, Is a permanent or on contract basis ,student is a regular student or part time. Student opted one or many subject. Whereas using semantic net we can't represent the knowledge scene wise. Semantic net can't be use to represent the knowledge event by event. So to get all the knowledge from the system, integrated knowledge representation technique is used. The hybrid structure is shown in fig 8. From script to semantic net two different directional link coming out that shows the link between the roles of script with the two different classes of semantic net. In the same way we can make the link between other roles and objects involve in scripts (scene wise) with the class and object in the semantic net. The unnamed link in semantic net shows the generalization for eg. Mode can be part time, full time and regular.

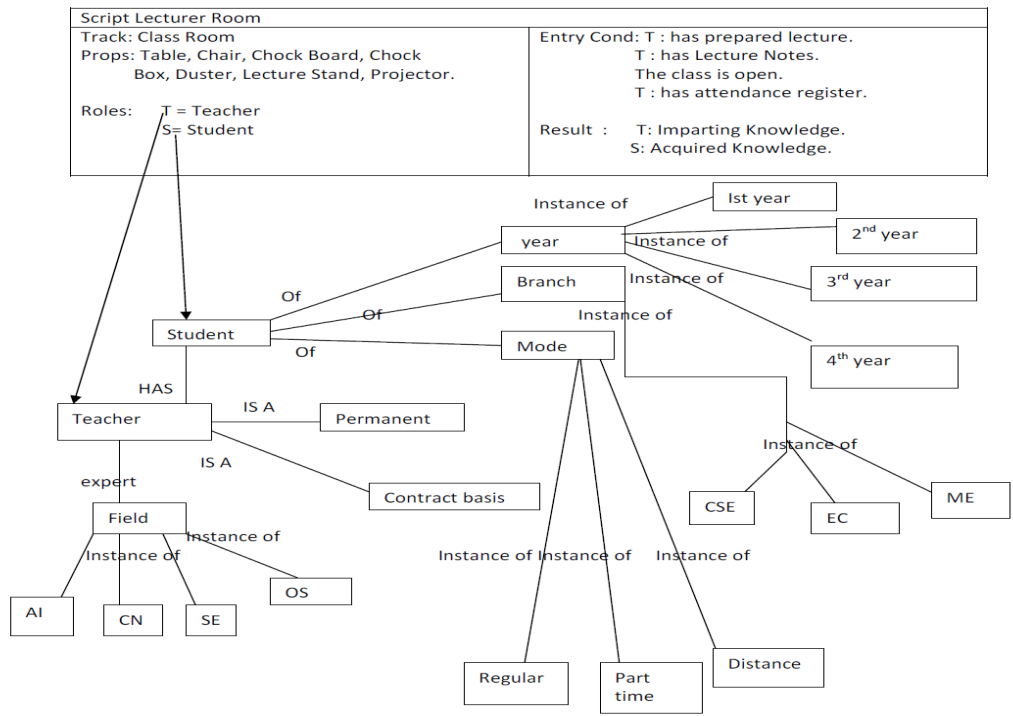

Fig 8. Hybrid Knowledge Representation technique.

As we know the input to the system is a sentence / paragraph/ story. Let us consider the another ex for representation.

Story 1. Ram Navami

Ram Navami is a Hindu festival that celebrates the birth of Lord Rama to king Dasharatha and queen Kaushalya of Ayodhya. Rama was an incarnation of Lord Vishnu. On this day, devotees of loard rama keep a fast. Houses are cleaned and temples are decorated. Offering of fruits and flowlers are made to the deity. It is customary to read massages from Ramayana.
The Hybrid representation for the above story is shown fig 9.

## 3.1 STRENGTH OF HYBRID KNOWLEDGE REPRESENTATION TECHNIQUE.

Human beings use past/previous learning & senses to understand verbal communication and in actual real world events do follow stereotyped patters. Communication style of each one is different from other and it is quite often when relating events, do leave large amount of blanks/gaps or assumed details out of their communication. This may lead to miscommunication. In real life it is not easy to deal with a system that is not able to fill up the missing conversational features, whereas scripts can predict/ assume unobserved events. Scripts can fill the gaps created from incomplete/disjoined observations and can build a sequential information. Semantic net is best knowledge representation technique for representing non event based knowledge with its technical simplicity. Even non technology savvy can also extract information/ knowledge from the semantic net.

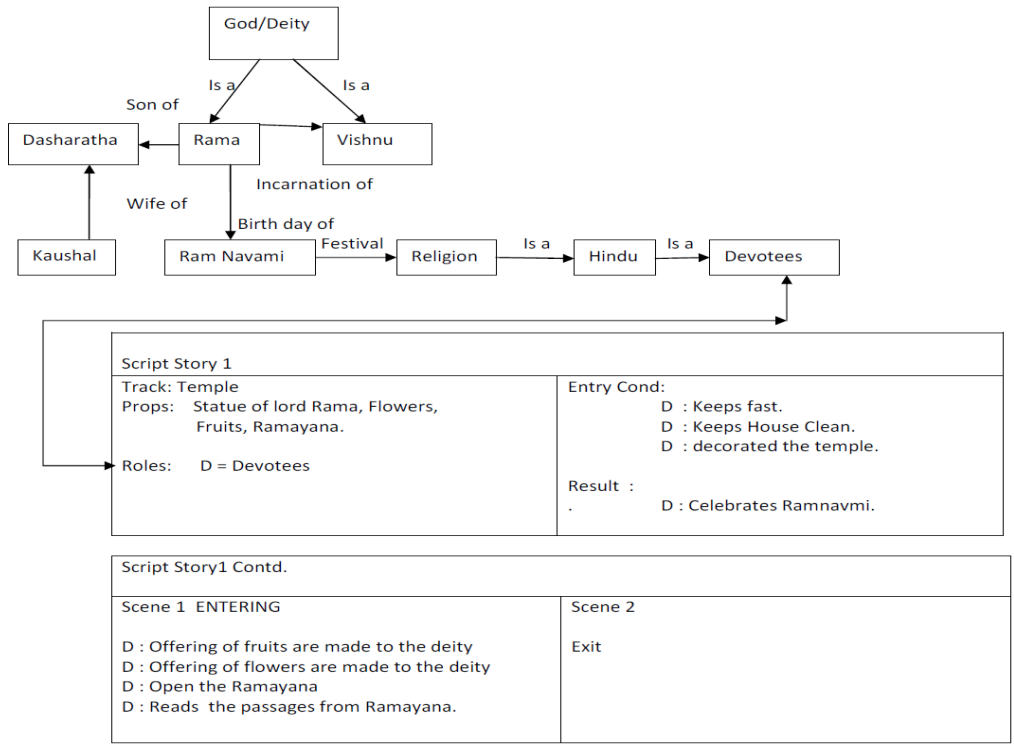

Fig 9. Hybrid representation of story 1.

## 4. CONCLUSION

There are various knowledge representation schemes in AI. All KR techniques have their own semantics, structure as well as different control mechanism and power. Combination of two or more representation scheme may be used for making the system more efficient and improving the knowledge representation. We are trying to build the intelligent system that can learn itself by the query and have a power full mechanism for representation and inference. The semantic net and script are very powerful techniques in some respects so the aim is to take the advantage of these techniques under one umbrella. The comparison between various hybrid KR techniques is shown in table with the proposed one.

Table Comparison between hybrid knowledge representation techniques

| S.no | Knowledge Structure/ properties | Year | Author | Technical description | Example | Diagram | Advantages | Disadvantges/Limitations | Applications | References |
|---|---|---|---|---|---|---|---|---|---|---|
| 1 | KRYPTON | 1983 | 1 Ronald J.Brachman  2 Richard E. Fikes  3 Hector J. Levesque | Two boxes are used terminological box(T box) and assertion box (A box). TBox used the structure of KL-ONE in which terms are organized taxonomically, using frames an ABox used the first-order logic sentences for those predicates come from the TBox, and a symbol table maintaining the names of the TBox terms so that a user can refer to them. It is basically a tell ask module. All interactions between a user and a KRYPTON knowledge base are mediated by TELL and ASK operations. | Ex of operations on Tbox  ABox:  TELL: KB X SENTENCE -> KB  Sentence is true.  ASK: KB x SENTENCE + {yes, no, unknown}  Is sentence true?  TBox: TELL:  KB X SYMBOL X TERM -> KB  By symbol, I mean term.  ASK: KB x TERM X TERM -> {yes, no}  Does a term1 subsume term2  ASK: KB x TERM x TERM → {yes, no}  Is term1 disjoint from term2 | 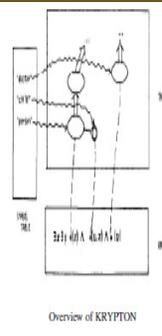  Overview of KRYPTON | Combine the advantages of both frame based and pradicate logic. Can work for incomplete information. | 1 The expressive power is limited  2 the limited interface between user and a knowledge base. the user is only limited to what is knowledge base rather than the functionality. | Natural language processing. | [14], [16] |
| 2 | Oblog-2 | 1987 | T. F Gordon | It combines a terminological reasoner with a Prolog(inference mechanism). The terminological component supports the description of type and attribute taxonomies. Entities are instances of a set of types. Horn clause rules are used as a Procedures for determining the values of attributes . The known types of an entity determine its set of applicable rules, which changes as our knowledge about the types of the entity is refined, supporting a form of defensible reasoning. Laws can be represented as general rules with exceptions, a technique traditionally used in the law, together with burden of proof rules, for reaching decisions when less than perfect information is available. | Kb clauses[ goal * knowledge base]->; Kb clauses( g, kb)=  If the Entity argument of g is bound then entity clauses(g,e), where e is the entity.  Else  Interleaved (entity – clauses(g,e1)...entity v(g, en)) where e1....en are the asserted entities of kb.  Entity clauses[ goal * entity] --> clauses; | 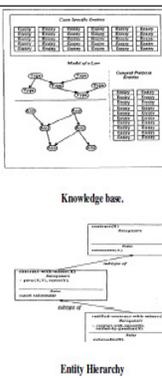  Knowledge base.  Entity Hierarchy | 1 Oblog-2 has been designed for modeling legal domains.  2 Can work for incomplete information . | 1 The control component has no memory of goals because the oblog inherits from its Prolog core. If any subgoal faces a problem during the solving of some larger problem, it begins again at the top of the rule hierarchy, instead of catching up the problem where it last left off.  2 In Krypton system can infer the relationships between two type by a direct comparison of the descriptions of the types. Where as in case of Oblog 2 it is not possible, i.e The relationship between two types in Oblog must be asserted. | 1 Initially used to build a variety of "toy" legal expert systems.  2 This system can be used in courses for law students on legal applications of AI.  3 Works for non monotonic reasoning. | [41] |

| S.no | Knowledge Structure/ properties | Year | Author | Technical description | Example | Diagram | Advantages | Disadvantges/Limitations | Applications | References |
|---|---|---|---|---|---|---|---|---|---|---|
| 3 | Babylon | 1987 | 1 Hans W.  2. Guesgen, Ulrich Junker,  3 Angi Voß, | It is the hybrid of Frame, rule and the prolog & incorporated with a system having the capability of constraint satisfaction. This version of Babylon can be used as a planning system, a knowledge acquisition system and a system for process diagnosis are being developed, all of which depend on the availability of some constraint mechanism. Babylon has a dynamic database containing frame instances, items produced by rules, and facts asserted by prolog. | This the Ex, for traffic light with two fires  defconstraint  (:name inverse- state)  (type primitive)  (interface fire 1 fire 2)  Relation(:tuple(on off) (:tuple (off on)))). | 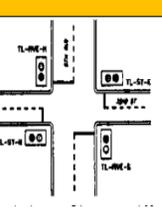  Represents Crossing between 5th avenue and 32nd street. | 1 Can be used for process diagnosis, planning. | Fully dependent on constraint. | Processing diagnosis & Planning. | [45],[46] |
| 4 | MANTRA | 1991 | 1 J. Calmet, I.A. Tjandra  2 G. Bittencourt | It is the combination of four different knowledge representation techniques. First order logic, terminological language , semantic networks and Production systems. all algorithm used for inference are decidable because this representation used the four valued logic. Mantra is a three layers architecture model. It consist the epistemological level, the logical level, Heuristic level. | Ex of operation in logic level  1 command ::= tell(knowledge base, fact) .  2 ask( knowledge base, Query)  3 to-frame(frame-def)  4 to-met( snet-den)  5 Fact ::= to-logic(formula)  6 Query ::= from logic(formula)  Ex of operation on terminological box  frame - def ::= identifier : c = concept  l identifier : r = relat ion  concept ::= ( concept ) l conce pt . | Architecture is defined in terms of operation and interfaces.  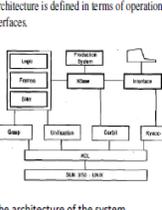  The architecture of the system. | 1 An intelligent, graphical user interface would help in building knowledge bases.  2 Expressive.  3 support procedural knowledge  4 A graph editor can be used to visualize, for instance, hierarchies or terminologies that would help the user for representing expert's knowledge by the use of | Less Expressive. | Only used for symbolic computation(Mathematical model) | [1] |

| S.no | Knowledge Structure/ properties | Year | Author | Technical description | Example | Diagram | Advantages | Disadvantges/Limitations | Applications | References |
|---|---|---|---|---|---|---|---|---|---|---|
| 5 | FRORL(frame- and-rule oriented requirement specification Language) | 1992 | 1 Jeffrey J. P. Tsai. 2 Thomas Weigert. 3 Hung-Chin Jang. | FRORL is based on the concepts of frames and production rules. Which is designed for software requirement and specification analysis. Six main steps are following, 1) Identify subject and themes. 2) Define object frames. 3) Define object abstract inheritance relation 4) Define object attributes. 5) Identify activity frames. 6) Define actions and communications There are two types of frames Object frame and activity frames Object Frames are used to represent the real world entity not limited to physical entity. These are act as a data structure. Each activity in FRORL are represented by activity frame to represent the changes in the world. Activity, Precondition and action are reserved word not to be used in specification. Language of FRORL consist of Horn clause of predicate logic. | Overview of software using FRORL | FRORL Software development methodology | 1 Modularity. 2 Incremental development. 3 Reusability. 4 Prototyping | Constraints must be defined properly. | Used for building prototype model for software. | [10][11] |
| 6 | TEX-I-bylon | 1993 | 1 Uwe Busbach 2 Rainer Jahnk. | TEX-I-bylon, which is based on the Babylon 2.1 shell. The main extension with respect to Babylon is the ability to have knowledge bases running concurrently. This goal has been achieved by mapping knowledge bases onto operating systems' processes. We will refer to these concurrent knowledge bases as inference tasks'. The benefits gained from the original Babylon are representation formalisms for frames, rules, constraints and Prolog terms. | 1 Find implication : car-control: do one rule as transaction( change – position). 2 Do one rule as transaction <rule-list> & optional <time out> | Represents two schemes for Pessimists currency control / Various usage of the transaction mechanism in rule set | 1 Used to run the knowledge bases in concurrent manner. | | 1 Used to run the knowledge bases in concurrent manner like operating system. | [42] |

| S.no | Knowledge Structure/ properties | Year | Author | Technical description | Example | Diagram | Advantages | Disadvantges/Limitations | Applications | References |
|---|---|---|---|---|---|---|---|---|---|---|
| 7 | AAANTS( Adapted autonomous , Agent colony interaction with Network Trans parent Services | 2003 | R.A. Chaminda Ranasinghe | AAANTS model was developed to built a prototype for an Intelligent Room that actively adapts the environmental conditions based on user behavioral patterns. AAANTS is a combination of frame based knowledge representation with reinforcement learning technique. AAANTS is a multi-agent system. frames were used to represents the different states of activation. Function is associated with each state. The value of this function return the expected future, relates to a value function that indicates the expected future. | Ex when an agent switch on the light it may nave following condition 1 For reading Book 2 some one came in the room 3 for searching any thing | Knowledge representation using AAANTS | 1 Dynamic 2 Expressive 3 supports multimedia | This system Can't work for incomplete information. | Intelligent environment. | [19] |
| 8 | Extended Semantic Net | 2009 | 1 Reena T. N. Shetty. 2 Pierre-Michel Riccio. 3 Joel Quinqueton | It is a hybrid of Proximal network model and Semantic network model. The proximity model is used to categorise the entities that relate to one another for interactive learning. When a number of entities are close in proximity a relationship is implied and if entities are logically positioned; they connect to form a structural hierarchy. this technique is used for to enable processing of large amount of data in short span of time. The proximal network model involves three phase of processing, 1 the documents related to the domain are analyzed called pre-treatment process  2  output of word document matrix is obtained. 3 This matrix is then passed on to the intermediate process and is analyzed by the data mining and clustering algorithms . K-means Clustering algorithm is used for this task. The semantic network consist of nodes and is based on the KLONE model, the domain components define the concepts using the instance and inheritance relations. | Extract of Extended Semantic visualization using graph editor | Extended semantic Network | 1 Expressive . 2 good balance between mind and mathematical models to develop better knowledge representing. | The system is not able to make the right balance for combining the concept network of semantic network with word network obtained from the proximal network. | Document classification and indexation. 2 The main application of ESN is in the area of environmental nuclear toxicology. | [43] |

| S.no | Knowledge Structure/ properties | Year | Author | Technical description | Example | Diagram | Advantages | Disadvantges/Limitations | Applications | References |
|---|---|---|---|---|---|---|---|---|---|---|
| 9 | Proposed system | - | Poonam Tanwar, Dr T.V Prasad, Dr Kamlesh Dutta, | The KR system must be able to represent any type of knowledge, "Syntactic, Semantic, logical, Understanding ill formed input, Ellipsis, Case Constraints, Vagueness". This system consist five main parts the K Box, Knowledge Base, Reasoning, Query applier and user interface. The Input from the user is in two forms either a new information or a query. If incoming input is the new information then it goes to the Acquisition and learning process to check the following knowledge is already in knowledge base if yes then skip. Otherwise checks whether that knowledge will be accommodate by the existing system if yes then follow the segmentation process. | | Represents the architecture of the system | 1 Ability to represents structural as well as non structural knowledge. 2 Dynamic Knowledge base 3 Ability to represents stereotype knowledge. 4 Strong user interface. | To be in process | Natural language processing. • Question answer system. | [39],[44] |

**Authors 1**

Poonam Tanwar received her B.Tech, M.Tech degree in Computer Science & Engg from Maharishi Dayanand University Haryana. Prsuing P.hd From Uttarakhand Technical University Dehradun( UTU). She has over 10 years of experience in teaching. Currently she is Assit Professor At Lingaya's University, Faridabad, Haryana, India. She has 20 papers to his credit. Her areas of Interest includes artificial intelligence, Computer graphics, Theory of computation, Soft Computing. She is member of IEEE.

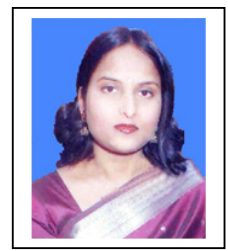

**Authors 2**

Dr. T. V. Prasad received his master's degree in Computer Science from Nagarjuna University, AP India and a doctoral degree from Jamia Milia Islamia University, New Delhi, India.. He has over 17 years of experience in industry and teaching. He has worked as Deputy Director in the Bureau of Indian Standards, New Delhi. Currently he is Professor and Dean academic. At Lingaya's University, Faridabad, Haryana, India. He has 100 papers and three books to his credit. His areas of interest include bioinformatics, artificial intelligence, consciousness studies, computer organization and architecture. He is a member of reputed bodies like ISRS, CSI, APBioNet, etc.

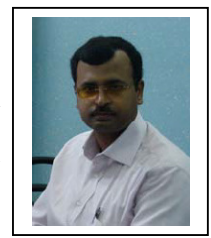

**Authors 3**

**Dr. Kamlesh Dutta** is working as Associate Professor in the Department of Computer Science & Engineering in National Institute of Technology, Hamirpur (India).